\documentclass[sigconf]{acmart}

\usepackage{algorithm}
\usepackage{algorithmic}

\usepackage{graphicx}
\usepackage{subfigure}
\usepackage{enumitem}
\usepackage{booktabs}
\usepackage{bbding}
\usepackage{color}
\usepackage[table,xcdraw]{xcolor}
\usepackage{multicol}
\usepackage{colortbl}
\usepackage{amsmath}
\usepackage{multirow}
\AtBeginDocument{%
  }

\newcommand{\M}{ContextRL}

\usepackage{tcolorbox}
\usepackage{listings}

\begin{document}

\title{\M{}: Enhancing MLLM's Knowledge Discovery Efficiency with Context-Augmented RL}

\author{Xingyu Lu$^{\dagger}$}
\affiliation{
  \institution{Tsinghua Shenzhen International Graduate School, Tsinghua University}
  \city{Shenzhen}
  \country{China}
}
\email{luxy22@mails.tsinghua.edu.cn}

\author{Jinpeng Wang$^{\ddag}$}
\affiliation{%
  \institution{Harbin Institute of Technology, Shenzhen}
  \city{Shenzhen}
  \country{China}
}
\email{wangjp26@gmail.com}

\author{Yi-Fan Zhang}
\affiliation{%
  \institution{Chinese Academy of Sciences}
  \city{Beijing}
  \country{China}
}
\email{yifanzhang.cs@gmail.com}
\author{Shijie Ma}
\affiliation{%
  \institution{Chinese Academy of Sciences}
  \city{Beijing}
  \country{China}
}
\email{mashijie2021@ia.ac.cn}

\author{Xiao Hu}
\affiliation{
  \institution{Tsinghua University}
  \city{Beijing}
  \country{China}
}
\email{hu-x21@mails.tsinghua.edu.cn}

\author{Tianke Zhang}
\affiliation{%
  \institution{Kuaishou Technology}
  \city{Beijing}
  \country{China}
}
\email{zhangtianke@kuaishou.com}

\author{Changyi Liu}
\affiliation{%
  \institution{Kuaishou Technology}
  \city{Beijing}
  \country{China}
}
\email{liuchangyi@kuaishou.com}
\author{Haojie Ding}
\affiliation{%
  \institution{Kuaishou Technology}
  \city{Beijing}
  \country{China}
}
\email{dinghaojie@kuaishou.com}

\author{Kaiyu Jiang}
\affiliation{%
  \institution{Kuaishou Technology}
  \city{Beijing}
  \country{China}
}
\email{jiangkaiyu@kuaishou.com}
\author{Kaiyu Tang}
\affiliation{%
  \institution{Kuaishou Technology}
  \city{Beijing}
  \country{China}
}
\email{tangkaiyu@kuaishou.com}
\author{Bin Wen$^{\star}$}
\affiliation{%
  \institution{Kuaishou Technology}
  \city{Beijing}
  \country{China}
}
\email{wenbin@kuaishou.com}

\author{Fan Yang}
\affiliation{%
  \institution{Kuaishou Technology}
  \city{Beijing}
  \country{China}
}
\email{yangfan@kuaishou.com}

\author{Tingting Gao}
\affiliation{%
  \institution{Kuaishou Technology}
  \city{Beijing}
  \country{China}
}
\email{lisize@kuaishou.com}

\author{Han Li}
\affiliation{%
  \institution{Kuaishou Technology}
  \city{Beijing}
  \country{China}
}
\email{lihan08@kuaishou.com}

\author{Chun Yuan$^{\ddag}$}
\affiliation{%
  \institution{Tsinghua Shenzhen International Graduate School, Tsinghua University}
  \city{Shenzhen}
  \country{China}
}
\email{yuanc@sz.tsinghua.edu.cn}

\thanks{$\dagger$ Work done during an internship at Kuaishou Technology.}
\thanks{$\ddag$ Corresponding Authors: Jinpeng Wang and Chun Yuan}
\thanks{$\star$ Project Leader.}
\renewcommand{\shortauthors}{Xingyu Lu et al.}

\begin{abstract}
Reinforcement Learning with Verifiers (RLVR) has become a standard paradigm for post-training Multimodal Large Language Models (MLLMs): During the sampling process, the policy model generates experience. The verifier then distinguishes this experience into correct knowledge and erroneous patterns, and subsequently guides the policy model to internalize the valid knowledge while avoiding the incorrect patterns, thereby improving overall model performance. Although effective, we argue that current RLVR frameworks suffer from two intrinsic information bottlenecks that hinder effective knowledge discovery: (1) Identifiability: Verifiers with limited context (e.g., only final answers) struggle to distinguish correct reasoning from hallucinations, leading to reward hacking; (2) Reachability: For hard queries, the policy model rarely samples correct responses, resulting in sparse gradient signals. In this work, we propose ContextRL, a novel framework that leverages context augmentation to overcome these bottlenecks.
Specifically, to enhance Identifiability, we provide the reward model with full reference solutions as context, enabling fine-grained process verification to filter out false positives (samples with the right answer but low-quality reasoning process). To improve Reachability, we introduce a multi-turn sampling strategy where the reward model generates mistake reports for failed attempts, guiding the policy to "recover" correct responses from previously all-negative groups.
Experimental results on 11 perception and reasoning benchmarks show that ContextRL significantly improves knowledge discovery efficiency. Notably, ContextRL enables the Qwen3-VL-8B model to achieve performance comparable to the 32B model, outperforming standard RLVR baselines by a large margin while effectively mitigating reward hacking. Our in-depth analysis reveals the significant potential of contextual information for improving reward model accuracy and document the widespread occurrence of reward hacking, offering valuable insights for future RLVR research.
\end{abstract}




\maketitle
\section{INTRODUCTION}

Knowledge discovery \cite{KD2} is defined as a systematic process to identify valid, novel, and interpretable patterns from large-scale data and transforming them into actionable knowledge structures \cite{KD}. In the big data era, how to perform knowledge discovery under with large-scale heterogeneous multimodal data becomes a core challenge in the construction of intelligent information systems.
In recent years, Multimodal Large Language Models \cite{qwen3vl, glmv,kimivl} (MLLMs) have demonstrated remarkable capabilities in multimodal understanding and reasoning tasks \cite{keye,o3}, offering critical support for knowledge acquisition \cite{visualKD} and representation \cite{LLMKG, LLMKG2}.

The training of MLLMs can be conceptualized as a parameter-centric process of knowledge discovery: models acquire knowledge from training datasets driven loss functions, with the learned representations encoded within parameters. Building upon Large Language Model (LLM)\cite{PLM, flamingo}, MLLM training is typically divided into two stages: pre-training and post-training \cite{qwen3vl,mimovl}. During pre-training, models leverage large-scale image–text corpora to achieve multimodal alignment \cite{qwen1,vita}. In post-training, common approaches include knowledge distillation via supervised fine-tuning (SFT) \cite{simpleo3, llavaov} and policy optimization through reinforcement learning (RL) \cite{RL,deepeyes}. For both pre-training and SFT, the model directly acquires knowledge from static datasets. In contrast, RL requires the MLLM to function as a policy model that interactively explores an environment composed of a reward system and a dataset, receiving feedback to iteratively refine its behavior, enabling MLLMs to acquire knowledge adaptively across diverse scenarios \cite{vla1, vla2}.

Inspired by large reasoning LLMs \cite{dsv3,reason}, RL training for MLLMs commonly adopts the RLVR (Reinforcement Learning with verifier Reward) paradigm \cite{visionr1}. In this framework, the MLLM samples from the training set to generate responses for each query, which are then evaluated by a Verifier that provides correctness-based feedback. The model subsequently adjusts the generation probabilities of different responses according to this feedback signal. RLVR-based methods, such as GRPO \cite{grpo} and DAPO \cite{dapo}, have demonstrated considerable success. Current optimization on RL algorithms primarily focuses on reward shaping \cite{quantile}, regularization \cite{entropy}, and training-inference consistency \cite{gspo}, these approaches fail to increase the amount of knowledge that the policy model acquires from the environment during RL, and thus do not overcome the information bottleneck of RLVR systems. 

To elucidate this, we conduct an in-depth analysis: We first identify two information sources for the policy model in RLVR framework: (1) responses generated by sampling, which constitute exploratory experience, and (2) reward signals provided by the verifier, which discriminate experience into knowledge and mistakes. We then delineate the information bottlenecks of each source: For experience sampling, if the policy model cannot produce correct responses, no actionable optimization signal is conveyed, thereby stalling learning; for knowledge discrimination, if the verifier does not provide accurate judgment, the model becomes susceptible to reward hacking \cite{hack1}. Both bottlenecks substantially constrain the knowledge discovery efficiency of MLLMs during RLVR training.

Recognizing these limitations, we propose leveraging context augmentation to overcome both information bottlenecks. We introduce the \textbf{ContextRL} framework, which consists of: (1) Context-Augmented Reward Model that enhances the reliability of reward signals by providing the reward model with richer reference context, thereby mitigating reward hacking caused by false-positive samples; (2) Context-Augmented Policy Model that employs a multi-turn sampling strategy, wherein mistake reports derived from negative responses are fed back to the policy model to expand its knowledge boundaries and facilitate the acquisition of correct responses; and (3) tailored optimization procedures for both single-turn and multi-turn samples to ensure training stability and convergence. 

To validate the efficacy of ContextRL in enhancing knowledge discovery for MLLMs, we constructed a training dataset comprising 29K samples to train the \texttt{Qwen3VL-8B-Instruct} \cite{qwen3vl} model. We conduct comprehensive evaluations on 5 perception benchmarks and 6 reasoning benchmarks, comparing ContextRL against SFT and other RLVR methods. Experimental results demonstrate that ContextRL successfully injects more reliable knowledge into \texttt{Qwen3VL-8B}, significantly outperforming baslines. Notably, the ContextRL-trained 8B model achieves performance comparable to that of the 32B variant.
To further elucidate the underlying mechanisms of ContextRL, we performed in-depth analytical experiments. Results indicate that context augmentation substantially improves the discriminative capability of the reward model. Moreover, we observed that false-positive samples are prevalent during training and pose a significant threat to effective knowledge acquisition by MLLMs. Finally, we quantitatively characterize the information gain introduced by ContextRL, providing empirical evidence for its effectiveness in alleviating information bottlenecks within the RLVR framework.

We summarize our contributions as follows:

\begin{itemize}
\item  \textbf{An in-depth analysis}: We figure out the information bottlenecks inherent in the RLVR framework, highlighting the critical importance of reward reliability and the generation of correct responses for effective knowledge acquisition in MLLMs.

\item \textbf{A novel and powerful framework}: We propose ContextRL, leveraging context augmentation to enhance both the policy model and the reward model. Evaluations across 11 diverse benchmarks demonstrate that ContextRL substantially improves the efficiency of knowledge discovery during RLVR training, expands the MLLM's knowledge boundaries, and exhibits strong generalizability.

\item \textbf{Insightful findings}: Through analytical experiments, we reveal the detrimental impact of false-positive samples on MLLM learning, uncover the substantial potential of context in improving reward model accuracy, and document the pervasive occurrence of reward hacking phenomena, providing valuable insights and practical guidance for future RLVR research.
\end{itemize}
\section{METHODOLOGY}

\subsection{RLVR for MLLMs}
\label{sec:rlvr_kd_prelim}

RLVR provides an efficient mechanism for MLLMs to extract, consolidate, and store knowledge from interactions with the environment.

\subsubsection{RLVR's Components}
A typical RLVR system for MLLMs consists of the following elements:

\textbf{Dataset:}
Let $\mathcal{D}$ denote a set of queries and corresponding ground truth,
$
\mathcal{D} = \{(x, t)\},
$
$x$ is a multimodal query (e.g., text-image input), and $t$ is $x$'s ground-truth annotation (e.g., answer or solution).

\textbf{Policy Model:}
The policy model $\pi_\theta$ is an MLLM parameterized by $\theta$.
Given a query $x$, $\pi_\theta$ can generate a response:
$
y \sim \pi_\theta(\cdot \mid x).
$

\textbf{Verifier:}
A verifier $V$ evaluates the quality (e.g., correctness) of a response by taking the query $x$, the generated response $y$, and the ground truth $t$ as inputs.
It outputs a scalar reward signal:
$
R(x,y) = V(x, y, t).
$
The verifier $V$ can be instantiated as rule-based programs, powerful MLLMs or human annotations.

\subsubsection{RLVR Workflow: GRPO as an Example}
We take the famous Group Relative Policy Optimization (GRPO) algorithm to illustrate RLVR's optimization dynamics.
Given a query-ground-truth pair $(x,t)\sim\mathcal{D}$, RLVR proceeds as follows.

\textit{(1) Experience Sampling.} The policy model samples a group of $G$ responses for $x$:
$
\{y_1, \ldots, y_G\} \sim \pi_\theta(\cdot \mid x).
$
In this step, the policy explores $x$'s response space to produce diverse experiences.

\textit{(2) Knowledge Discrimination.}
The verifier assigns a reward score to each sampled response to discriminate between good and bad experiences:
$
\{R_1, \ldots, R_G\},  R_k = R(x, y_k) = V(x, y_k, t).
$

\paragraph{(3) Advantage Construction.}
To obtain relative advantages within the sampled group, rewards are normalized:
$
A(x,y_k) = \frac{R(x,y_k) - \mu_x}{\sigma_x},
$ where $\mu_x$ and $\sigma_x$ denote the mean and standard deviation of $\{R_1,\ldots,R_G\}$

\paragraph{(4) Knowledge Internalization.}
Finally, the policy model updates parameters to increase the generation probability of responses with a positive advantage and decrease the negatives' generation probability.
A standard policy-gradient estimator is formulated as:
\begin{equation}
\nabla_\theta J(\theta) \approx 
\mathbb{E}_{x \sim \mathcal{D},\, y \sim \pi_\theta(\cdot \mid x)}
\left[ A(x, y)\, \nabla_\theta \log \pi_\theta(y \mid x) \right].
\label{eq:pg}
\end{equation}
Through repeated updates, the policy model internalizes verifier-endorsed patterns as implicit knowledge encoded in $\theta$.

\subsubsection{Brief Summary}
Overall, RLVR enables the policy model to explore candidate behaviors on multimodal queries, leverages a verifier to distinguish high-quality experiences from low-quality ones, and consolidates the discovered knowledge by shifting generation probabilities via advantage-weighted optimization.

\subsection{Information Bottlenecks in RLVR}
\label{sec:rlvr_bottleneck}

Despite the success of RLVR in enhancing the internal knowledge and capabilities of MLLMs, we argue that the standard RLVR paradigm suffers from intrinsic \emph{information bottlenecks}.
As formalized in Section~\ref{sec:rlvr_kd_prelim}, the knowledge that the policy model $\pi_\theta$ can internalize is determined by:
(i) whether $\pi_\theta$ can \emph{reach} informative (ideally correct) solutions during experience generation, and
(ii) whether the verifier can reliably \emph{identify} correctness  during experience discrimination.
Accordingly, RLVR can bottleneck at either stage.

\subsubsection{Bottleneck I: Reachability of Positive Solutions}
\label{sec:bottleneck_reachability}

RLVR relies on policy sampling to produce high-quality experiences.
However, when the probability of sampling a correct response is small, the learning signal becomes extremely sparse.
Let $C \in \{0,1\}$ denote the \emph{true correctness} of a sampled response $y$ for query $x$, where $C=1$ indicates a positive solution (e.g., correct answer with valid reasoning).
Define the \emph{per-sample success probability} under the current policy as
$
p_x \triangleq \Pr_{y \sim \pi_\theta(\cdot \mid x)}[C=1].
$
In group-based RLVR, the probability of obtaining \emph{at least one} positive solution in the group is
$
q_x(G) \triangleq 1 - (1 - p_x)^G.
$
For hard queries, $p_x$ may be extremely small, yielding $q_x(K) \approx 0$ even for moderately large $G$.
Consequently, sampled groups are often \emph{all-negative}, i.e., $C(y_k)=0$ for all $k$.
This regime causes a fundamental bandwidth collapse of the learning signal.
Intuitively, the policy receives predominantly ``what not to do'' feedback, but rarely observes ``what to do'' exemplars.

\subsubsection{Bottleneck II: Identifiability of Correctness}
\label{sec:bottleneck_identifiability}

Even if the policy successfully samples candidate solutions, RLVR further depends on the verifier to correctly discriminate good/bad experiences.
We highlight that this discrimination can be inherently ambiguous when the verifier operates under \emph{restricted context}:
Let $T$ denote the context available to the verifier when assessing a response $y$ for query $x$.
In typical RLVR practice, the ground-truth signal is often minimal (e.g., only a final answer $t$). In such settings, negative samples that do not match $t$ can be easy to reject, but \emph{false positives} may arise: a response may match the final answer while being incorrect in the reasoning process. Formally, identifiability requires that correctness be determined by the verifier's context, i.e., $C$ is a deterministic function of $T$.
When this is not the case, there exists irreducible uncertainty captured by the conditional entropy
\begin{equation}
H(C \mid T) > 0.
\label{eq:cond_entropy_bottleneck}
\end{equation}
When Eq.~\eqref{eq:cond_entropy_bottleneck} holds, even an optimal verifier cannot perfectly infer the true correctness label under the limited context, and the scalar reward becomes potentially biased and noisy.
This bias directly contaminates the advantage signals and reduces training efficiency.
\begin{figure*}[t]
    \centering
    \includegraphics[width=\textwidth]{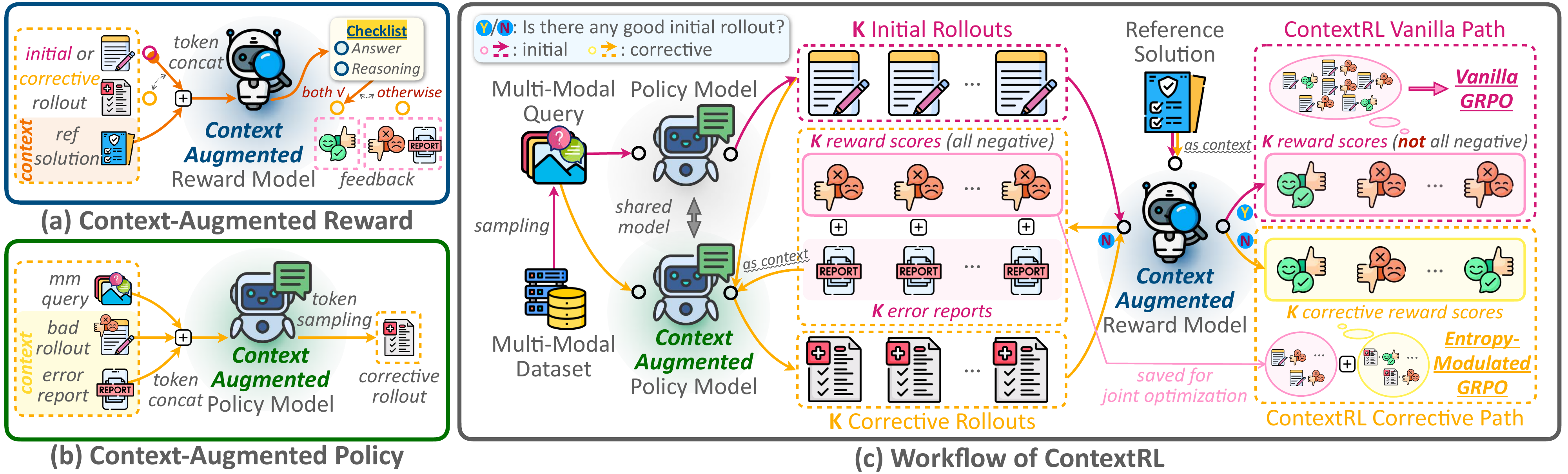}
    \caption{Overview of ContextRL. (a) \textbf{Context-augmented reward model}. (b) \textbf{Context-augmented policy.} (c) \textbf{Training workflow.}}
    \label{fig:stage2}       
    \Description{method figure}
\end{figure*}
\section{ContextRL: Augmenting RLVR with Context}
\label{sec:method}

To address the information bottlenecks of RLVR, we propose \textbf{ContextRL} to augment both the policy model and the reward model with additional context. (1) For the reward model, we provide a {full solution} rather than a minimal final answer, enabling fine-grained process evaluation to reduce false positives. (2) For the policy model, we provide {mistake reports} and conduct a second-stage generation  to increase the reachability of correct solutions

\subsection{Context-Augmented Reward Model}
\label{sec:context_augmented_verifier}

\paragraph{Reward Context: Full Solution vs. Final Answer.}
Compared to regular ground truth with only the final answer $a$, we introduce the \emph{full solution} $s$ for each query $x$ to enrich the reward context, which includes both the reasoning process and the final answer for $x$:
\begin{equation*}
T_0 = (x, y, a)\ \text{(regular RLVR)}, 
\qquad 
T_1 = (x, y, s)\ \text{(ContextRL)}.
\end{equation*}
Here, $a = g(s)$ is the final answer in $s$. The augmented context $T_1$ provides more comprehensive information compared to $T_0$, which allows the verifier to perform a finer-grained assessment of the response $y$ by comparing $y$ against $s$ in detail. This results in a reduction of the uncertainty $H(C \mid T_1)$, improving the identifiability of correctness compared to $T_0$:
$
H(C \mid T_1) \le H(C \mid T_0),
$
with strict inequality when the correctness of the response depends on intermediate steps present in $s$ but absent from $a$.

\paragraph{Context-Augmented Reward Model.}
Given the augmented reward context $(x,y,s)$, the reward model now produces two outputs: (i) a scalar reward $R(x, y)$ for evaluating the correctness of $y$, and (ii) a mistake report $M(x, y^-)$ for each negative sample $y^-$, which identifies errors in $y^-$. These outputs are computed as:
\begin{equation*}
R(x, y) = r_\psi(x, y; s), \quad M(x, y^-) = h_\psi(x, y^-; s).
\end{equation*}
The mistake report $M(x, y^-)$, generated with respect to the full solution $s$, explicitly pinpoints the issues in $y^-$.

\subsection{Context-Augmented Policy Model}
\label{sec:context_augmented_policy}

\paragraph{Stage-1: Standard Group Sampling.}
In stage-1, the policy model samples $G$ responses $\{y_g\}_{g=1}^G \sim \pi_\theta(\cdot \mid x)$ for each query $x$. These responses are then evaluated by the context-augmented verifier to obtain rewards and mistake reports for negative responses. If at least one positive response is found, the sampling process terminates.

\paragraph{Stage-2: Context-Augmented Sampling.}
If stage-1 receives an all-negative sample group, we proceed to conduct stage-2 sampling: We append the query $x$, each negative response $y_g^-$, and its mistake report $M_g$ into $\pi_\theta$'s context and initiate a second-round generation:
\[
\{y_g^{(2)}\}_{g=1}^G \sim \pi_\theta(\cdot \mid x, y_g^-, M_g).
\]
This context-augmented generation process guides $\pi_\theta$ to avoid previous issues and increase the chances of sampling positive solutions.

\paragraph{Sample Filtering.}
After stage-2, we filter the samples to retain only those that meet two criteria: (i) \textbf{correctness}, as verified by the reward model, and (ii) \textbf{independence}, meaning the stage-2 sample does not mention the previous negative response or mistake report. Let $\mathcal{P}(x)$ represent the set of retained positive samples:
\[
\mathcal{P}(x) = \left\{ y_i^{(2)} : R_\psi(x, y_i^{(2)}; s) = 1 \text{ and } y_i^{(2)} \text{ is independent} \right\}.
\]

\subsection{ContextRL's Optimization Process}
\label{sec:method_opt}

ContextRL uses two types of training groups for optimization.

\paragraph{Online Training Group.}
For queries that receive at least one positive response in stage-1, stage-1 samples produce an online training group for $\pi_\theta$ (online means no other conditions):
\[
G_{\mathrm{on}}(x) = \{(x, y_k^{(1)})\}_{k=1}^{K}, \qquad \exists k : R(x, y_k^{(1)}) = 1.
\]
This group undergoes training using the standard GRPO objective, with group-relative advantages computed over all samples.

\paragraph{Mixed Training Group.}
In cases where the stage-1 group contains no positive samples and stage-2 sampling provides high-quality positives, we form a mixed group consisting of both online negatives and offline positives (with $y^-$ and $M$ as condition). Specifically, for each stage-2 positive $y \in \mathcal{P}(x)$, we apply context rollback to  remove  $y^-$ and $M$ and convert it into a single-turn sample $(x, y^{(2)})$. The mixed training group is defined as:
\[
G_m = \underbrace{\{(x, y_k)\}_{k=1}^K}_{\text{online negatives}} \cup \underbrace{\{(x, y) : y \in \mathcal{P}(x)\}}_{\text{offline positives}}.
\]
The group undergoes optimization using the GRPO objective with scaled advantages for the mixed group and selective KL regularization to handle the offline-stitched positives.

\paragraph{Advantage Scaling.}
To mitigate the impact of the mixed training group on policy entropy, we scale the advantages for mixed groups:
\[
\widetilde{A}(x, y) = \lambda A(x, y), \qquad \lambda \in (0, 1),
\] $\lambda$ is a hyperparameter to controls the influence of mixed groups.

\subsection{ContextRL Algorithm}
\label{sec:method:algorithm}
Pseudocode~\ref{alg:contextrl} describes the full ContextRL pipeline.

\begin{algorithm}[h]
\caption{ContextRL Pipeline}
\label{alg:contextrl}
\begin{algorithmic}[1]
\REQUIRE Policy $\pi_\theta$; reward model $(r_\psi)$; dataset $\mathcal{D}$ providing $(x,s)$
\FOR{each query $x \sim \mathcal{D}$}
    \STATE \textbf{Stage 1 (standard group sampling):} $\{y_k^{(1)}\}_{k=1}^K \sim \pi_\theta(\cdot\mid x)$
    \STATE \textbf{Produce rewards and mistake report:}
    \STATE \hspace{1.5em} $R_k^{(1)} = r_\psi(x,y_k; s), \quad M^{-} = h_\psi(x,y^{-}; s)$
    \IF{ $\exists k$ s.t.\ $R_k^{(1)}=1$ }
        \STATE \textbf{Online group update:} compute advantages on $\{R_1^{(k)}\}$ and update $\pi_\theta$ with GRPO objective
    \ELSE
        \STATE \textbf{Stage 2 (context-augmented sampling):} sample $\{y_k^{(2)}\}_{k=1}^K \sim \pi_\theta(\cdot\mid x,y^-,M)$
        \STATE \textbf{Filter standalone positives:} $\mathcal{P}(x)=\Big\{y_i^{(2)}:{R}(x,y_i^{(2)};s)=1\ \ \text{and}\ \ y_i^{(2)}\ \text{is independent}\Big\}.$
        \STATE \textbf{Context rollback:} form stitched pairs $\{(x,y): y\in\mathcal{P}(x)\}$ and mixed training group $G_x$
        \STATE \textbf{Mixed group update:} compute group-relative advantages on $G_x$, scale by $\lambda$, update $\pi_\theta$
    \ENDIF
\ENDFOR
\end{algorithmic}
\end{algorithm}

\section{EXPERIMENTS}
\label{sec: experiments}
\subsection{Experimental Setup}
\subsubsection{Dataset Construction}
To construct a training dataset with Reference Solution, we select FineVision \cite{finevision} as the data source. The Policy Model used for optimization is Qwen3-VL 8B Instruct, while the Reward Model is Qwen3-VL 32B Instruct. To validate the generality of ContextRL over tasks, we choose VQA and multimodal math problems to conduct experiments. We select several VQA dataset (e.g., ArXivQA \cite{arxivqa}, ThinkLite \cite{thinklite}, AI2D \cite{ai2d}) and several Math dataset (e.g., MMK12 \cite{mmk12}, GEOQA \cite{geoqa}) from FineVision. The candidate dataset consists of approximately 250,000 data instances each with query and final answer. We then filter all instances based on accuracy \cite{prime}: For each instance, four responses are generated by the Policy Model, and the Reward Model evaluates their correctness according to the ground truth. Instances with all correct responses are filtered out. The remaining instances are then re-sampled using the stronger Reward Model to generate positive samples. From these positive samples, the Reward Model selects the optimal solution as the reference solution for enhancing reward modeling context. To balance the difficulty, the data is categorized based on the number of correct responses generated by the Policy Model, ensuring an equal number of samples in each category. Ultimately, the training dataset consists of  16K VQA instances and approximately 14K multimodal math instances, each containing a multimodal query, a reference solution, and its final answer.

\subsubsection{Benchmarks}
To comprehensively evaluate the effect of ContextRL in enhancing the perception and reasoning capabilities of MLLMs, we select the corresponding two types of benchmarks. For perception evaluation, we choose five benchmarks: SimpleVQA \cite{simplevqa}, MMStar \cite{mmstar}, HallusionBench \cite{hallusionbench}, HRBench8K \cite{hr8k}, and MME-RealWorld-lite \cite{MMERealWorld}. MMStar  primarily evaluates the general capabilities of MLLMs across multiple dimensions, HallusionBench and SimpleVQA focus on assessing hallucinations in generated outputs, while HRBench8K and MME-RealWorld-Lite examine the performance of MLLMs in high-resolution scenarios. For reasoning evaluation, we select six benchmarks: MathVerse \cite{mathverse}, MathVista \cite{mathvista}, LogicVista \cite{logicvista}, We-Math \cite{wemath}, CharXiv-RQ \cite{charxiv}, and DynaMath \cite{dynamath}. MathVerse, MathVista, and We-Math target visual mathematical reasoning, emphasizing fine-grained understanding and multi-step reasoning, LogicVista focuses on visually grounded logical reasoning, systematically assessing models’ abilities in inductive, deductive, spatial, and symbolic reasoning tasks. CharXiv-RQ evaluates scientific chart reasoning, and DynaMath tests MLLMs' consistency under dynamic visual and textual scenarios.

\subsubsection{Baselines} 
To validate the effectiveness of ContextRL in enhancing policy model's knowledge compared with SFT paradigm and other RLVR methods, we adopt the following baselines:

(1) SFT: We directly fine-tune the policy model using the reference solutions as training targets. This process is equivalent to distilling knowledge from the reward model, which acts as a teacher, combined with rejection sampling on the policy model.

(2) GRPO \cite{grpo}: As introduced earlier (Section~\ref{sec:rw_rl_vlm}), GRPO is the most commonly applied RLVR approach, which improves the accuracy of the policy model by computing intra-group relative advantages and applying policy gradient optimization.

(3) DAPO \cite{dapo}: DAPO is a well-known variant of GRPO that enhances training stability and learning efficiency through multiple techniques including clip-shifting, dynamic sampling, token-level policy gradient Loss, and overflowing reward shaping.

(4) Qwen3-VL 32B Instruct: Qwen3-VL 32B Instruct is the largest dense model in the Qwen3-VL open-source family and significantly outperforms Qwen3-VL 8B Instruct across a wide range of tasks.

\subsubsection{Implementation Details}
For all our training experiments, we adopt the \texttt{QwenVL-3-8B-Instruct} model as the policy model and \texttt{QwenVL-3-32B-Instruct} as the reward model, with their maximum pixel budgets set to 3M and maximum response lengths capped at 16k tokens. During SFT, the model is trained on 29K samples for 5 epochs, with a learning rate of 1e-5 and a global batch size of 128.

For ContextRL and GRPO, the KL divergence loss coefficient $\beta$ is set to 0.01. The global batch size is configured to 128, the sampling group size to 8, and the learning rate to 1e-6. All reinforcement learning experiments are conducted for a single training epoch. In both SFT and RL phases, we employ the AdamW optimizer \cite{adamw} with a \texttt{cosine\_with\_min\_lr} learning rate scheduler and a warm-up ratio of 0.03. All SFT and RL training is performed on 8 NVIDIA H800 GPUs using the ms-swift framework \cite{swift}.

All the evaluation is carried out via VLMEvalKit framework, with GPT-4o \cite{4o} judging model generated solutions' correctness. For SFT, we select the checkpoint exhibiting the highest average performance across the 5 training epochs; for RL methods, we directly report results from the single-epoch checkpoint.

\begin{table*}[ht]
    \centering
    \caption{Performance comparison of different models on perception and reasoning Benchmarks. We report the evaluation metrics on each split of the benchmark to provide a fine-grained presentation of the models' performance. Optimal
and sub-optimal performance for each metric is denoted in bold and underlined fonts, respectively.}
    \resizebox{0.95\textwidth}{!}{
    \begin{tabular}{lccccccccccc}
        \toprule 
        \rowcolor{black!7}\multicolumn{12}{c}{\emph{{Perception Benchmarks}}} \\ 
        \toprule
        {\textbf{Benchmark}} & {\textbf{SimpleVQA}} & {\textbf{MMStar}} & \multicolumn{3}{c}{\textbf{HallusionBench}} &  \multicolumn{3}{c}{\textbf{HRBench8K}} & \multicolumn{2}{c}{\textbf{MME-Real-lite}} & \multirow{2}{*}{\textbf{Avg.}} \\ \cline{2-3}\cline{3-4}\cline{4-7}\cline{7-10}\cline{10-11}
        {\emph{split}}& \emph{overall} & \emph{overall} & \emph{aAcc} & \emph{fAcc} & \emph{qAcc} & \emph{single} & \emph{cross} & \emph{overall} & \emph{percept} & \emph{reason} \\ 
        \midrule 
        Qwen3-VL 8B & 44.93 & 64.06 & 71.08 & 53.46 & 53.19 & 79.00 & 61.00 & 70.00 & 54.66 & 41.73 & 59.31 \\ 
        SFT & 46.14 & 70.87 & 74.97 & 56.36 & \underline{56.48} & 67.75 & 70.00 & 71.88 & 55.95 & 50.27 & 62.07 \\ 
        GRPO & 46.86 & 71.87 & \underline{75.50} & 58.09 & 55.60 & 80.50 & 70.25 & 75.38 & 57.83 & 50.67 & 64.25 \\ 
        DAPO & 47.62 & 72.00 & 75.29 & \textbf{60.98} & 54.73 & \textbf{82.25} & 68.75 & \underline{75.50} & \textbf{58.51} & \underline{51.73} & \underline{64.74} \\ 
        ContextRL & \underline{48.35} & \underline{73.00} & \textbf{75.71} & \underline{60.40} & \textbf{56.48} & \underline{80.75} & \textbf{70.50} & \textbf{75.63} & \underline{58.17} & \textbf{53.20} & \textbf{65.22} \\ 
        \midrule 
        Qwen3-VL 32B & \textbf{53.07} & \textbf{74.53}  & 75.08 & 55.49 & 55.38 & 79.25 & \underline{70.50} & 74.88 & 57.14 & 45.73 & 64.10 \\ 
        \bottomrule
        \toprule
        \rowcolor{black!7}\multicolumn{12}{c}{\emph{{Reasoning Benchmarks}}} \\         \toprule
        {\textbf{Benchmark}}  & \multicolumn{1}{c}{\textbf{MathVerse}} & \multicolumn{1}{c}{\textbf{MathVista}} & \small{\textbf{LogicVista}} &  \multicolumn{2}{c}{\textbf{We-Math}} & \multicolumn{3}{c}{\textbf{CharXiv-RQ}} & \multicolumn{2}{c}{\textbf{DynaMath}}& \multirow{2}{*}{\textbf{Avg.}} \\ \cline{2-11}
       {\emph{split}}  & \emph{mini} & \emph{mini} & \emph{overall} & \emph{strict} & \emph{loose} & \emph{chart} & \emph{general} & \emph{overall} & \emph{worst} & \emph{overall} \\ 
        \midrule 

        Qwen3-VL 8B & 60.50 & 76.80 & 56.82 & 55.33 & 72.67 & 50.36 & 45.68 & 46.00 & 39.52 & 67.49 &57.12 \\ 
        SFT & 64.82 & 76.80 & 59.06& 55.43  & 73.52 & 48.38 & 47.95 & 47.80& 38.32 & 67.37 & 57.94 \\ 
        GRPO & 64.64 & 78.80 & 57.04 & 61.62 & 75.90 & 47.32 & 43.66 & 44.90 & 40.11 & 67.80 & 58.17 \\ 
        DAPO & 65.17 & 77.70 & \underline{60.85} & \underline{63.81} & \underline{78.67} & 51.29 & 51.13 & 48.90 & 41.17 & 68.08 & 60.68 \\ 
        ContextRL & \underline{69.34} & \underline{78.90} & 60.40 & \textbf{64.48} & \textbf{81.24} & \underline{51.75} & \underline{51.23} & \underline{50.40} & \underline{47.52} & \underline{68.42} & \underline{62.37} \\ 
        \midrule 
        Qwen3-VL 32B & \textbf{69.40} & \textbf{82.40} & \textbf{62.20} & {63.52} & 78.00 & \textbf{55.06} & \textbf{58.24} & \textbf{55.20} & \textbf{49.70} & \textbf{76.53 }& \textbf{65.03} \\ 
        \bottomrule
    \end{tabular}
    }
    \label{tab:performance}
\end{table*}

\subsection{Main Results}
In {Table}~\ref{tab:performance}, we present the performance of {ContextRL} and other baselines on five perception benchmarks and six reasoning benchmarks. From these results, we draw the following conclusions:

(1) \textbf{{ContextRL} yields significantly larger performance gains.}
    Across all perception benchmarks, {ContextRL} achieves the best or second-best results. On most reasoning benchmarks, it attains the second-best performance, only behind Qwen3-VL 32B Instruct, while on We-Math, {ContextRL} outperforms Qwen3-VL 32B Instruct. Moreover, {ContextRL} consistently surpasses SFT and the other two RLVR methods. We attribute this advantage to {ContextRL}'s ability to break the information bottleneck inherent in RLVR systems.

(2) \textbf{{ContextRL} demonstrates generality across tasks.}
    {ContextRL} delivers consistent performance improvements on both perception and reasoning tasks. On perception benchmarks, it improves the average performance of Qwen3-VL 8B by 5.91\%, while on reasoning benchmarks, it achieves a 5.25\% gain. In contrast, SFT, GRPO, and DAPO tend to yield larger improvements on simpler perception tasks, whereas {ContextRL} exhibits comparable gains across both task categories, validating its generality. We attribute this to {ContextRL}'s use of multi-round sampling to enhance the reachability of positive samples for hard instances, thereby providing richer reward signals in challenging reasoning tasks.

(3) \textbf{{ContextRL} is robust to the quality of reference solutions.}
    We observe that directly applying SFT with reference solutions can result in unchanged or even degraded performance on certain benchmarks (e.g., HRBench8K, MathVistaMini, and DynaMath). This suggests potential dataset bias across benchmarks, as well as suboptimal quality in some reference solutions generated by the reward or policy models. In contrast, by incorporating reference solutions as contextual information for reward modeling, {ContextRL} achieves stable performance improvements, indicating strong robustness to the quality of reference solutions.

\textbf{(4) Divergence between perception and reasoning tasks.} Improving Qwen3-VL 8B's performance on perception tasks is easier compared to more complex reasoning tasks. The Qwen3-VL 8B model trained with {ContextRL} surpasses the Qwen3-VL 32B on perception tasks, but still lags behind on reasoning tasks. The  difficulty of reasoning tasks results in a larger performance gap between the initial 8B and 32B models, and  RL training brings less performance improvement on reasoning benchmarks. We attribute this to the small scale of mathematical instances in our training data. We are working to conduct experiments with more reasoning training data to achieve more significant improvements.

\subsection{Analysis Experiments}
\subsubsection{Context-augmented reward modeling significantly improves error detection.}
\label{sec:fp}
We first evaluate the reliability of our context-augmented reward model in identifying `false positives': responses that arrive at the correct final answer through flawed reasoning. To establish an evaluation dataset for analysis, we collect responses rejected by our ContextRL reward model. We cross-verify these negatives with \texttt{Qwen3-VL-235B-Insturct} and retain only those consistently classified as negative by both models, labeling them as \textbf{High-Confidence False Positives}. To validate this automated classification, we manually inspected a random subset of 1,000 such samples. Human verification confirmed that 95.8\% of these samples indeed contained reasoning errors or hallucinations. And we take 30k false-positive responses for the following experiment.

To validate the effective of context augmentation for detecting false positives, we investigate the impact of different reference information in the context on reward modeling. We select Qwen3-VL 32B and Qwen3-VL 235B as reward models and test them under three context settings: (1) no reference provided ({w/o ref}), (2) final answer as reference ({w. answer}), and (3) full solution as reference {w. solution}. The models are required to judge the correctness of the high-confidence false-positive samples. We report the ratio of samples that are identified as wrong by the reward models in Table~\ref{tab:reward_model_comparison}.

The results show that the proportion of false-positive samples identified as errors increases with the amount of reference information. 
The higher identification rate supports our hypothesis that additional reference information results in a reduction of uncertainty ($H(C \mid T)$). 
We further observe that, as the reference information increases, the performance gap between the 32B and 235B reward models gradually narrows. 
This indicates that, given sufficient in-context reference information, smaller models can achieve reward accuracy comparable to that of larger models. This holds significance for developing of efficient MLLM RL system.

\begin{table}[h]
\centering
\caption{Reward Accuracy Analysis: The proportion of identified high-confidence false-positive samples under different reward models across context settings.}
\resizebox{0.4\textwidth}{!}{
\begin{tabular}{lccc}
\hline
Reward Model & \small{\textbf{w/o ref}} & \small{\textbf{w. answer}} & \small{\textbf{w. solution}} \\
\hline
\small{Qwen3-VL 32B}  & 46.25 & 50.03 & 81.98 \\
\small{Qwen3-VL 235B} & 51.25 & 54.60 & 82.12 \\
\hline
\end{tabular}
}
\label{tab:reward_model_comparison}
\end{table}
\begin{table}[h]
\centering
\caption{Model performance after SFT with different proportions of false-positive samples in the training data.}
\resizebox{0.5\textwidth}{!}{
\begin{tabular}{lcccccc}
\hline
\textbf{\small{Benchmark}} & \textbf{\small{MME-Real}} & \textbf{\small{MMStar}} & \textbf{\small{HRBench8K}} & \textbf{\small{MathVista}} & \textbf{\small{We-Math}} \\
\cline{2-6}
   {\emph{split}}  & \emph{lite} & \emph{overall} & \emph{overall} & \emph{mini} & \emph{strict} \\
\hline
\small{Qwen3-VL 8B}   & \large{49.61} &  \large{64.06} & \large{70.00} & \large{76.80} & \large{59.14} \\
all positive & \large{53.93} & \large{72.87} & \large{73.25} & \large{78.60} & \large{63.05} \\
w 10\% false & \large{53.88} & \large{73.93} & \large{73.00} & \large{78.30} & \large{62.76} \\
w 20\% false & \large{54.72} & \large{73.00} & \large{73.50} & \large{78.30} & \large{61.33} \\
w 30\% false & \large{53.41} & \large{72.47} & \large{73.49} & \large{77.00} & \large{61.29} \\
\hline
\end{tabular}
}
\label{tab:benchmark_negative_ratio}
\end{table}

\subsubsection{The influence of false-positive samples} To demonstrate the impact of reducing false-positive samples on model training, we conduct SFT experiments: 
We mix the false-positive samples with positive samples generated during training to construct 50k training datasets containing different proportions of false positives (ranging from 0\% to 30\%). 
We train the policy model using different versions of the datasets and examine the models' performance.

Table~\ref{tab:benchmark_negative_ratio} shows that training data consisting entirely of positive samples yields the largest performance gains. 
As the false-positive samples increases, model performance degrades on most benchmarks, with pronounced declines on reasoning benchmarks, MathVista and We-Math. This indicates that eliminating false-positive samples is of significant importance for improving the model’s knowledge discovery efficiency and ability.
However, there are also cases (e.g., HRBench8K) where the impact of false-positive samples is limited or even slightly beneficial. 
This phenomenon is related to task difficulty and evaluation protocols, as false-positive issues may also exist in the evaluation process (e.g., multiple-choice questions where correctness is determined by the final option).

The SFT experiments further indicate that positive samples generated by the policy model itself lead to better learning outcomes than directly using reference answers as training targets. 
We attribute this advantage of policy-generated positive samples to three main factors. 
First, they undergo stricter correctness evaluation than the reference solutions provided as in-context guidance. 
Second, they are more diverse: a single query can yield multiple positive samples, whereas only one reference solution is available. 
Third, they exhibit closer proximity to the policy model, making the self-generated samples easier for the model to learn from.

\subsubsection{Quantizing the information gain of ContextRL}
In the Appendix~\ref{sec:gain}, we quantize the information gain introduced by ContextRL from:
(1) the context-augmented reward model corrects the rewards of false-positive samples;
(2) context-augmented policy sampling provides non-zero advantages and gradients for all-negative groups.
Accordingly, the information gain of ContextRL is formulated as
$
\mathcal{I}_{\mathrm{ContextRL}}
=
\rho_{fp}
+
\rho_r .
$
Here, $\rho_{fp}$ denotes the fraction of false positives eliminated by the context-augmented reward system, and $\rho_r$ denotes the fraction of queries with positive samples recovered from stage-2 sampling among all queries in the training process. 

Using the strategy for identifying false positives (Sec~\ref{sec:fp}), we statistic the proportion of false positive samples in a single training epoch of ContextRL. An epoch of ContextRL generates approximately 230K responses (29K queries * 8 responses per query), of which 51.12\% are judged as negative by the context-augmented reward model, totaling 118K. Among these samples, we identify 20.6K false-positive samples, which account for 8.9\% of the total samples. During the training process, 8.56\% of the queries obtain positive samples from the stage-2 sampling. Considering that eliminating false positive samples would increase the proportion of all-negative groups in stage-1, the information gain of ContextRL compared to regular RLVR is approximately 17\%.

Our statistic results indicate a significant \textbf{reward hacking} issue in the traditional RLVR training process, where the policy model's reasoning contains mistakes, yet it produces answers that are consistent with the ground truth. Furthermore, compared to DAPO, which directly discards the all-negative group, ContextRL recovers these discarded samples by providing positive samples for difficult queries, resulting in a greater information gain.

\begin{table*}[ht]
    \centering
    \caption{Ablation Study:In the table, {CAR} denotes {context-augmented reward}. 
{CAS} denotes {context-augmented sampling}. 
{w/o report} indicates that the mistake report is not provided, and 
{w/o scale} indicates that advantage scaling is not applied.
}
    \resizebox{0.9\textwidth}{!}{
    \begin{tabular}{lccccccccc}
        \toprule
        \textbf{Method} & \textbf{\small{SimpleVQA}} & \textbf{MMStar} & \textbf{\small{HallusionBench}} & \textbf{\small{HRBench8K}} & \textbf{DynaMath} & \textbf{\small{MathVista}} & \textbf{\small{LogicVista}} & \textbf{We-Math} & \textbf{Avg.} \\
        \midrule
        ContextRL   & \textbf{48.35} & \underline{73.00} & \textbf{64.20} & \underline{75.63} & 68.42 & \textbf{78.90} & \textbf{60.40} & \textbf{64.48} & \textbf{66.67} \\
        \midrule
        w/o CAR     & 47.02 & 72.13 & 63.13 & 73.00 & \underline{69.64} & 78.60 & 55.93 & 58.19 & 64.71 \\
        w/o CAS   & 46.68 & 71.27 & 62.90 & 74.38 & {69.58} & 77.90 & 55.48 & 62.86 & 65.13 \\
        w/o report & 46.70 & \textbf{73.07} & 63.29 & 74.63 & \textbf{70.16} & 78.50 & \underline{60.36} & \underline{64.10} & \underline{66.35} \\
        w/o scale  & \underline{47.15} & 72.80 & 63.64 & \textbf{78.25} & 68.22 & \underline{78.80} & 59.13 & 61.90 & 66.24 \\
        \bottomrule
    \end{tabular}
    }
    \label{tab:ablation_study}
\end{table*}

\begin{figure*}[h]
    \centering
    \includegraphics[width=0.9\textwidth]{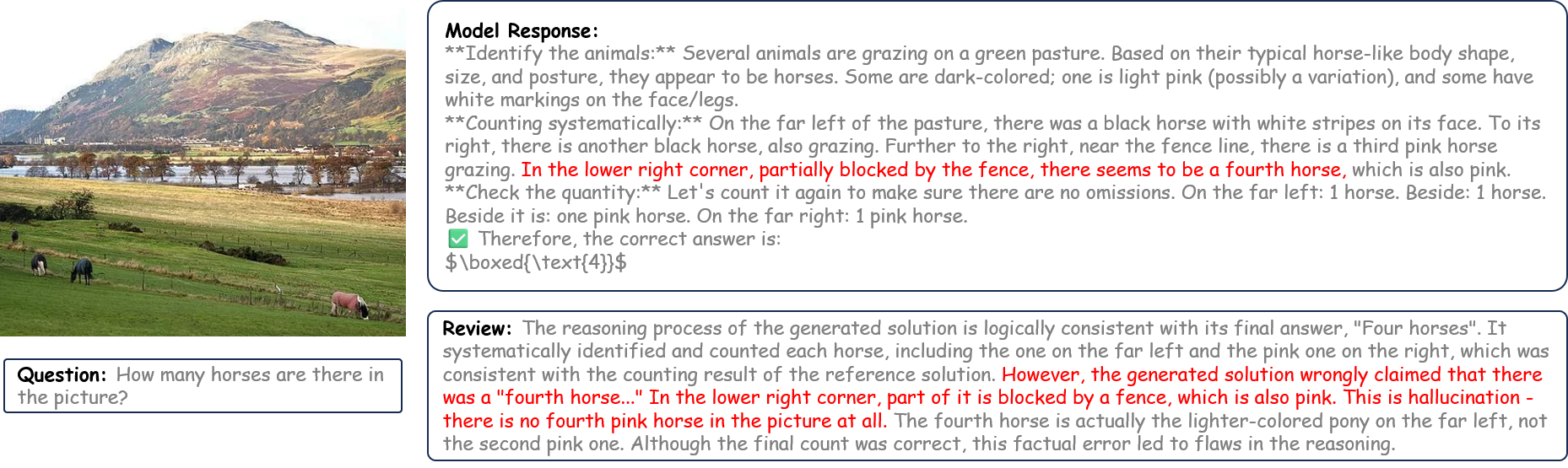} 
    \caption{\textbf{False Positive Example (Hallucination).}}
    \label{fig:fp_horse}
\end{figure*}

\subsection{Ablation Study}
In Table~\ref{tab:ablation_study}, we present a performance comparison between ContextRL and its four variants:
(1) \textit{without context-augmented reward model}: we replace the full solution in the reward model's context with the final answer;
(2) \textit{without context-augmented policy sampling}: we remove the multi-stage sampling procedure of ContextRL and perform only stage-1 sampling;
(3) \textit{without mistake report}: during stage-2 sampling, we do not provide the policy model with explicit mistake reports, and instead directly ask it to correct the erroneous stage-1 responses;
(4) \textit{without advantage scaling for mixed training groups}: for mixed groups containing stage-2 positive samples, we calculate loss using the unscaled GRPO advantage.

Our ablation study shows that the reference solution in the reward model's context brings the largest performance gain to ContextRL, further highlighting the importance of eliminating false positive samples and avoiding reward hacking in RL training. Removing multi-turn sampling and directly discarding hard queries also leads to performance degradation, which is particularly pronounced on reasoning benchmarks, indicating that providing positive samples for difficult data helps the policy model overcome its knowledge limitations. Moreover, providing no mistake report reduces the policy model's accuracy on hard samples, thereby decreasing the information gain. Finally, advantage scaling for mixed training groups contributes to improved training stability by reducing the influence of offline positive samples.

\subsection{Reward Hacking Case Study}

As shown in Figure~\ref{fig:fp_horse}, the model is tasked with counting horses in a landscape. While the final count ($4$) matches the ground truth, the intermediate reasoning reveals a severe hallucination. The model fails to identify a clearly visible horse on the far left but compensates for this count by fabricating a non-existent ``fourth horse'' in the lower right corner, describing it as ``partially blocked by the fence.'' This suggests the model may be gaming the counting objective by generating plausible-sounding descriptions for features that do not exist, rather than performing genuine visual grounding.

\section{RELATED WORKS}
\label{sec:related_work}

\subsection{MLLMs and Knowledge Discovery}
\label{sec:rw_vlm_kd}

Multi-Modal Large Language Models (MLLMs) have undergone rapid evolution in both architecture and training paradigms, progressing from early dual-encoder alignment models to modern MLLMs with strong generative reasoning abilities.
Early work typically learned joint visual--text representations via contrastive objectives, enabling robust cross-modal retrieval and transferable visual semantics (e.g., CLIP-style pretraining)~\cite{clip,zhai2023sigmoid}.
More recent MLLMs integrate a vision encoder with an auto-regressive language model, often through learnable adapters or projectors, and exhibit emergent capabilities in multimodal understanding, grounded generation, and multi-step reasoning~\cite{kimivl,keye,qwen3vl}.

This line of progress is closely related to \emph{knowledge discovery} \cite{KD}.
First, MLLM representations can be viewed as implicitly discovering structured regularities from large-scale multimodal corpora and storing them in parameters.
Such implicit knowledge is later \emph{extracted} through text generation (or probing), making MLLMs both repositories and interfaces of multimodal knowledge \cite{LLMKG4,LLMKG5}.
Second, MLLMs have also been explicitly connected to knowledge graphs (KGs) and knowledge discovery tasks \cite{LLMKG, LLMKG2, LLMKG3}.
Multimodal encoders provide unified embeddings for entities and relations grounded in both textual and visual evidence, which benefits KG completion and link prediction under sparse or ambiguous descriptions~\cite{chen2024knowledge}.

From a training perspective, contemporary MLLMs pipelines typically follow a two-stage recipe \cite{vita,qwen3vl}.
In the \emph{pretraining} stage~\cite{llava,qwen1}, models ingest broad and noisy web-scale data to acquire general world knowledge and aligned cross-modal representations.
In the \emph{post-training} stage \cite{mimovl,kimivl}, models improve knowledge utilization and controllability by increasing supervision quality and signal richness, including curated instruction tuning, preference optimization, and alignment objectives.
This paper follows this trend and focuses on improving the \emph{efficiency of knowledge discovery during post-training}, especially when reinforcement learning is used to internalize verifier-endorsed knowledge.

\subsection{Reinforcement Learning for MLLMs}
\label{sec:rw_rl_vlm}

Inspired by the success of reinforcement learning (RL) in aligning and enhancing reasoning in large language models\cite{dsv3,o3}, RL has become an important tool for improving MLLMs' perception \cite{thyme} and reasoning synergy \cite{keye, kimivl} and decision-making under interaction \cite{vla1, vla2}.
Compared to supervised fine-tuning \cite{llava,llavaov}, RL optimizes behavior by exploring the response space and exploiting evaluative feedback, which is widely believed to offer stronger generalization \cite{visionr1,deepeyes} when the supervision is weak, sparse, or underspecified.

A common practice for MLLM post-training is \emph{reinforcement learning with verifiers} (RLVR), where the model samples candidate responses and a verifier (rule-based, model-based, or human) assigns rewards.
Recent variants further improve stability and sample efficiency via group-based sampling and relative advantage estimation \cite{quantile} (e.g., GRPO-style updates) or via stronger regularization \cite{dapo} and preference-based objectives \cite{kuaimod,selfreward}.
Nevertheless, prior studies have pointed out that RLVR can be limited by the quality and informativeness of the verification signal \cite{yueyang}. 
When the verifier only checks a {final answer}, optimization may overfit to superficial cues and can fail to encourage correct intermediate reasoning, leading to reward hacking or spurious ``correct'' solutions.

In this work, we emphasize that RLVR may suffer from intrinsic \emph{information bottlenecks} that hinder knowledge discovery.
(i) \textbf{Reachability bottleneck:} for hard queries, correct solutions may be rarely sampled, making learning signals extremely sparse . 
Existing solutions include increasing the number of samples, curriculum design \cite{visionr1}, self-training/rejection sampling \cite{ragen}, or using auxiliary search/tool feedback \cite{minio3,pyvision}, but these approaches can be compute-intensive or task-specific.
(ii) \textbf{Identifiability bottleneck:} correctness judgment may be fundamentally ambiguous under restricted verification context, resulting in false positives and reward hacking \cite{hack2}.
Related directions attempt to provide denser supervision by incorporating process-level signals \cite{prime}, critique models \cite{judgerlvr}, reflection \cite{selfreward}, or step-wise rewards \cite{rstar}.

\section{CONCLUSION}
In this paper, we propose \textbf{ContextRL}, a novel framework to enhance the knowledge discovery efficiency of MLLMs during RL. ContextRL leverages context augmentation to overcome the information bottleneck in conventional RLVR methods.
Specifically, (1) ContextRL provides richer reference information to the reward model, improving reward accuracy and mitigating reward hacking; (2) ContextRL supplies explicit mistake reports, increasing the probability that the policy model generates positive samples for hard queries, thereby pushing the upper bound of MLLM knowledge capacity.
Experimental results on 11 benchmarks demonstrate that ContextRL consistently outperforms other RLVR methods. Notably, it enables {Qwen3-VL 8B} to achieve performance comparable to {Qwen3-VL 32B}, substantially improving the knowledge capability of MLLMs.
Further analytical experiments confirm the detrimental impact of reward hacking caused by false-positive samples and validate the effectiveness of incorporating contextual reference information to enhance reward model accuracy. We carefully quantify the information gain introduced by ContextRL and conduct ablation studies to verify the effectiveness of each component. Overall, ContextRL highlights the critical role of context information for broadening MLLM's knowledge boundaries.

\bibliographystyle{ACM-Reference-Format}
\bibliography{refs}

\clearpage
\appendix

\setcounter{section}{0}
\setcounter{figure}{0}

\renewcommand\thesection{\Alph{section}}
\renewcommand {\thefigure} {A-\arabic{figure}}

\renewcommand\thesubsection{\Alph{section}.\arabic{subsection}}

\section{APPENDIX}

\subsection{Quantifying the Information Gain of ContextRL}
\label{sec:gain}

In this section, we provide a quantitative analysis of the information gains introduced by ContextRL.
We show that ContextRL increases the effective optimization signal through two mechanisms:
(i) augmenting reward model's fidelity to reduce false positives, and
(ii) converting ineffective all-negative samples into informative training signals via context-augmented sampling.
We formalize both effects under a unified notion of \emph{information gain}.

\subsubsection{Information Gain from Reduced False Positives in Reward Modeling}
\label{sec:gain:rm}

Let $C\in\{0,1\}$ denote the true correctness of a sampled response and $\widehat{C}$ the binary decision induced by the reward model.
Under standard RLVR with minimal reference context, the reward system typically exhibits a non-negligible false positive rate:
\begin{equation}
\alpha_0 = \Pr[\widehat{C}=1 \mid C=0].
\end{equation}
Such false positives introduce spurious positive advantages, corrupting the policy-gradient signal.

With context-augmented reward modeling, the false positive rate is reduced to $\alpha_1 < \alpha_0$ due to improved identifiability.
Let
$
\rho_{\mathrm{fp}} = \alpha_0 - \alpha_1
$
denote the fraction of false positives eliminated  context-augmented reward system.
We define the reward-model information gain as
\begin{equation}
\mathcal{I}_{\mathrm{RM}}
=
\rho_{\mathrm{fp}},
\label{eq:gain_rm}
\end{equation}
which measures the proportion of previously misleading positive signals that are corrected and converted into reliable supervision.
This term directly reflects the increase in usable label information provided to the policy optimizer.

\subsubsection{Information Gain from Context-Augmented Sampling on All-Negative Queries}
\label{sec:gain:sampling}

In GRPO-style RLVR, queries whose sampled group is all-negative yield zero group-relative advantages and therefore contribute no effective gradient signal.
ContextRL introduces a context-augmented stage-2 sampling procedure that recovers correct solutions for a subset of these queries.
Let
$
\rho_r
$
denote the fraction of queries with stage-2 recovered positives among all queries in the overall training process.

For each group with $G$ stage-1 negatives, stage-2 positives transform these zero-advantage samples into effective policy-gradient contributors.
The information gain from context-augmented sampling is
\begin{equation}
\mathcal{I}_{\mathrm{TS}}
=
\rho_r.
\label{eq:gain_ts}
\end{equation}

\subsubsection{Unified Information Gain Expression}
\label{sec:gain:total}

Combining the two effects, we define the total information gain introduced by ContextRL as
\begin{equation}
\boxed{
\mathcal{I}_{\mathrm{ContextRL}}
=
\underbrace{\rho_{\mathrm{fp}}}_{\text{reward modeling gain}}
+
\underbrace{\rho_r}_{\text{two-stage sampling gain}}
}
\label{eq:gain_total}
\end{equation}

In particular, ContextRL (i) suppresses misleading gradients caused by false positives, and (ii) activates previously dormant queries by converting all-negative samples into effective learning opportunities.

\subsection{Instructions for Different Tasks}

In Figure~\ref{fig:regular_reward} and ~\ref{fig:context_reward}, we present two types of instructions used to guide our reward model in assessing the correctness of generated samples and in providing error reports. Figure~\ref{fig:regular_reward} illustrates the instructions used in regular RLVR, where the final answer is taken as the reference information. Figure~\ref{fig:context_reward} shows the context-augmented instructions employed in ContextRL, in which the full solution is used as the reference information.

\begin{figure}[!h]
\centering
\begin{tcolorbox}[
  colback=gray!10,colframe=black,
  title={Regular Reward Instruction},
  width=\linewidth,boxrule=0.8pt,arc=2mm,
  fonttitle=\bfseries,left=4mm,right=4mm,top=2mm,bottom=2mm,verbatim
]
You are an expert in visual question answering. Given a Question, a Reference Answer, and a Generated Solution (with \verb|`|reasoning\verb|`| and a \verb|`|final answer\verb|`| in \verb|\boxed{}|), evaluate:
(1) Self-Consistency; (2) Correctness vs. the image and Reference Answer; if Correctness=0, write a Mistake Report.

1. **Self-Consistency**:\\
   If the reasoning aligns with the final answer, score 1; otherwise score 1.\\
2. **Correctness** (only if Self-Consistency=1):\\
   If no hallucinations/conflicts with the image, reasoning is logically correct, and the final answer matches the Reference Answer in meaning, score 1; else 0.\\
3. **Mistake Report** (only if Self-Consistency=1 and Correctness=0):\\
   Summarize wrong content concisely; do not provide correct facts or reveal the correct final answer.\\

Inputs:\\
\#\#\#\# **Question**: \texttt{\textless Input Question\textgreater}\\
\#\#\#\# **Reference Answer**: \texttt{\textless Ref Answer\textgreater}\\
\#\#\#\# **Generated Solution**: \texttt{\textless Gen Solution\textgreater}\\
\#\#\# Output Format (JSON):
\begin{lstlisting}
{
  "Analysis": brief analysis of self-consistency and correctness,
  "Self-Consistency": score1,
  "Correctness": score2,
  "Mistake Report": "mistakes (if applicable)"
}
\end{lstlisting}
Your Evaluation Result:
\end{tcolorbox}
\Description{regular reward instruction}
\caption{Regular reward instruction template.}
\label{fig:regular_reward}
\end{figure}

\clearpage
\begin{figure}[!h]
\centering
\begin{tcolorbox}[
  colback=gray!10,colframe=black,
  title={Context-Augmented Reward Instruction},
  width=\linewidth,boxrule=0.8pt,arc=2mm,
  fonttitle=\bfseries,left=4mm,right=4mm,top=2mm,bottom=2mm,verbatim
]
You are an expert in visual question answering. Given a Question, a Reference Solution, and a Generated Solution (with \verb|`|reasoning\verb|`| and a \verb|`|final answer\verb|`| in \verb|\boxed{}|), evaluate Self-Consistency and then Correctness. If Correctness=0, provide a Mistake Report.\\
1. **Self-Consistency:**\\
   If pleasantries or revision cues (e.g., "You are absolutely right", "Thank you for your review", "I apologize for my mistake", "re-analyze", "reconstruct", "re-examine") appear, set Self-Consistency=0 and Correctness=0; else Self-Consistency=1 iff reasoning matches final answer (otherwise set both to 0).\\
2. **Correctness** (only if Self-Consistency=1):\\
   Using the image as truth and the Reference Solution as auxiliary, set Correctness=1 iff no hallucinations/conflicts, reasoning is logical, and the final answer matches the Reference Solution's final answer in meaning; else 0.\\
3. **Mistake Report** (only if Self-Consistency=1 and Correctness=0):\\
   List wrong reasoning content only; **no** evidence, correct facts, or final-answer leakage.\\

Inputs:\\
\#\#\#\# **Question**: \texttt{\textless Input Question\textgreater}\\
\#\#\#\# **Reference Solution**: \texttt{\textless Ref Solution\textgreater}\\
\#\#\#\# **Generated Solution**: \texttt{\textless Gen Solution\textgreater}\\
\#\#\# Output Format (JSON):
\begin{lstlisting}
{
  "Analysis": brief analysis of self-consistency and correctness,
  "Self-Consistency": score1,
  "Correctness": score2,
  "Mistake Report": "mistakes (if applicable)"
}
\end{lstlisting}
Your Evaluation Result:
\end{tcolorbox}
\Description{context-augmented reward instruction}
\caption{Context-augmented reward instruction template.}
\label{fig:context_reward}
\end{figure}

\subsection{Another Case}
\label{sec:reward_hacking}

To further investigate the limitations of outcome-based supervision, we visualize representative ``false positive'' cases. These are instances where the model arrives at the correct final answer (e.g., the correct count or multiple-choice option) but relies on flawed reasoning or hallucinations. In standard reinforcement learning settings that rely solely on outcome verifiers, these responses would receive a positive reward, encouraging the model to reinforce erroneous logic—a phenomenon known as reward hacking.

\begin{figure*}[h]
    \centering
    \includegraphics[width=\textwidth]{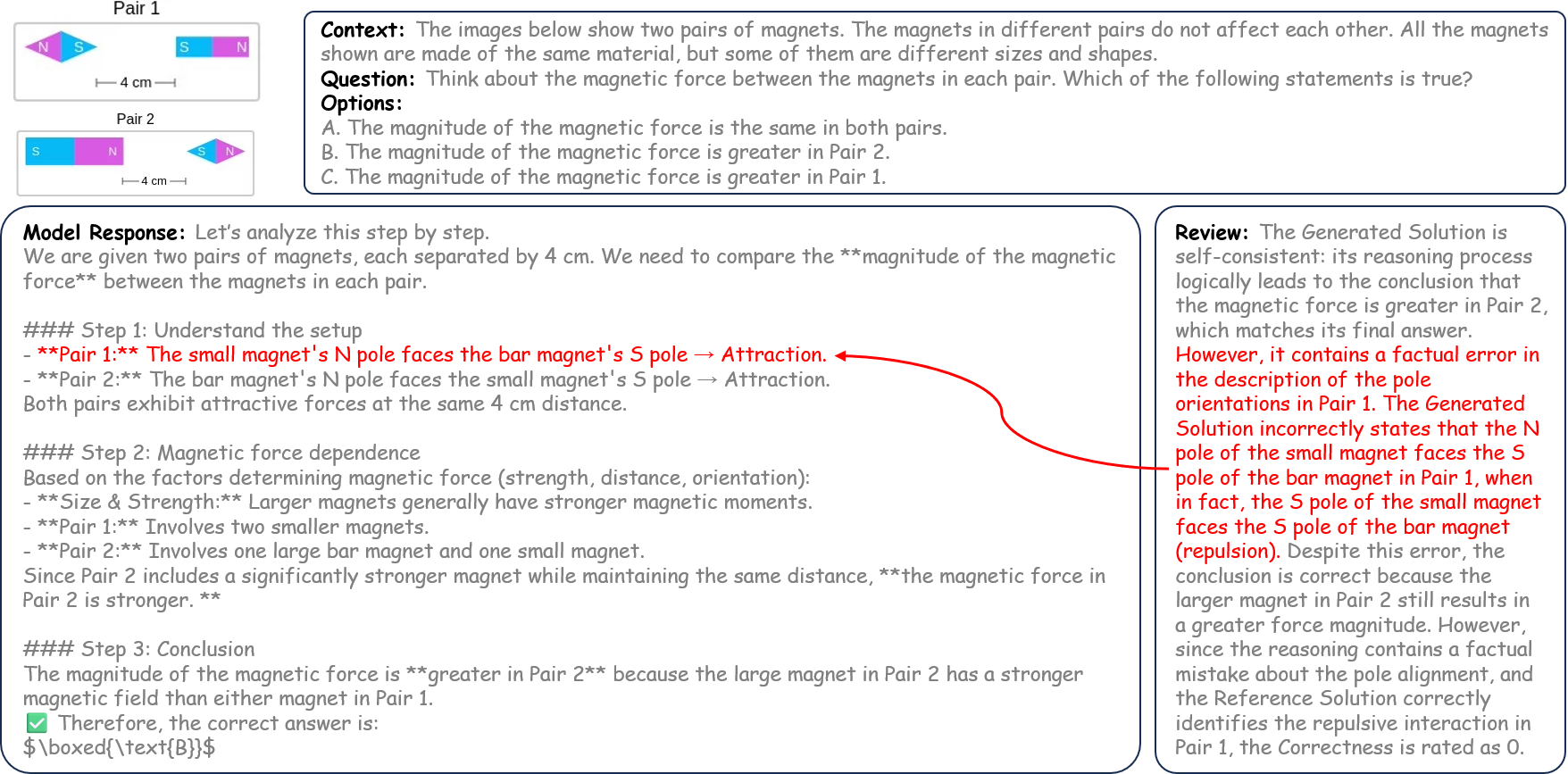} 
    \caption{\textbf{False Positive Example (Reasoning Error).}}
    \label{fig:fp_magnet}
\end{figure*}

\paragraph{Logical Inconsistency.} 
Figure~\ref{fig:fp_magnet} demonstrates a reasoning error in a physics problem regarding magnetic forces. The model correctly selects Option B; however, its derivation contains a fundamental factual error. It incorrectly analyzes the pole orientation in the first pair of magnets as ``attraction'' (North facing South) when the visual evidence clearly shows ``repulsion'' (South facing South). The correct final answer is reached only serendipitously. 

These cases highlight the necessity of the proposed review mechanism. As illustrated in the red text within the figures, our approach successfully detects these reasoning flaws, assigning a negative quality score despite the correct final outcome, thereby preventing the policy from learning these spurious correlations.

\end{document}